\documentclass[runningheads]{llncs}

\usepackage[T1]{fontenc}
\usepackage{graphicx}
\usepackage{comment}
\usepackage{amsmath,amssymb}
\usepackage{color}
\usepackage{url}
\usepackage{hyperref}

\usepackage{multirow}
\usepackage{algorithm}
\usepackage{algpseudocode}
\usepackage{mathabx}
\usepackage{booktabs}
\usepackage{array}
\usepackage{adjustbox}

\newif\ifreview
\reviewfalse

\ifreview
	\usepackage{lineno}

	\linenumbers
\fi

\begin{document}

%%%%%%%%%%%%%%%%%%%%% Add submission id, track, and title. %%%%%%%%%%%%%%%%%%%%%

% TODO: Please insert your submission number here
\def\SubNumber{090}

% TODO: Please uncomment the track this paper will be submitted to, comment all other lines
%\def\GCPRTrack{Main Track}
%\def\GCPRTrack{Special Track: Pattern recognition in the life and natural sciences}
%\def\GCPRTrack{Special Track: Photogrammetry and remote sensing}
%\def\GCPRTrack{Young Researcher's Forum}
\def\GCPRTrack{Fast Review Track}

% TODO: Replace with your title
\title{Drawing the Same Bounding Box Twice? Coping Noisy Annotations in Object Detection with Repeated Labels}
% You can use \thanks for acknowledgment as in: 
%\title{Title\thanks{XXX}}
%Do not add any acknowledgment to the draft 
% version that is used for the review process.  

\ifreview
	% ANONYMOUS SUBMISSION FOR REVIEW
	% DO NOT MODIFY these for the draft version that is used for the review process.
	\titlerunning{GCPR 2023 Submission \SubNumber{}. CONFIDENTIAL REVIEW COPY.}
	\authorrunning{GCPR 2023 Submission \SubNumber{}. CONFIDENTIAL REVIEW COPY.}
	\author{GCPR 2023 - \GCPRTrack{}}
	\institute{Paper ID \SubNumber}
\else
	\titlerunning{Coping Noisy Annotations in Object Detection with Repeated Labels}

	\author{David Tschirschwitz\orcidID{0000-0001-5344-4172} \and \\
	Christian Benz\orcidID{0000-0001-9915-0057} \and \\
        Morris Florek\orcidID{0009-0008-8425-5161} \and \\
        Henrik Norderhus\orcidID{0009-0006-2613-2572} \and \\
	Benno Stein\orcidID{0000-0001-9033-2217} \and \\
	Volker Rodehorst\orcidID{0000-0002-4815-0118}}
	
	\authorrunning{D. Tschirschwitz et al.}
	
	\institute{Bauhaus-Universität Weimar, Germany \\
	\email{david.tschirschwitz@uni-weimar.de}\\
 }
\fi

\maketitle              % typeset the header of the contribution

\begin{abstract}
The reliability of supervised machine learning systems depends on the accuracy and availability of ground truth labels. However, the process of human annotation, being prone to error, introduces the potential for noisy labels, which can impede the practicality of these systems. While training with noisy labels is a significant consideration, the reliability of test data is also crucial to ascertain the dependability of the results. A common approach to addressing this issue is repeated labeling, where multiple annotators label the same example, and their labels are combined to provide a better estimate of the true label. In this paper, we propose a novel localization algorithm that adapts well-established ground truth estimation methods for object detection and instance segmentation tasks. The key innovation of our method lies in its ability to transform combined localization and classification tasks into classification-only problems, thus enabling the application of techniques such as Expectation-Maximization (EM) or Majority Voting (MJV). Although our main focus is the aggregation of unique ground truth for test data, our algorithm also shows superior performance during training on the TexBiG dataset, surpassing both noisy label training and label aggregation using Weighted Boxes Fusion (WBF). Our experiments indicate that the benefits of repeated labels emerge under specific dataset and annotation configurations. The key factors appear to be (1) dataset complexity, the (2) annotator consistency, and (3) the given annotation budget constraints.

\end{abstract}
\section{Introduction}

\newcommand{\etal}{et. al.}

\begin{figure}[ht]
\begin{center}
\includegraphics[width=1.0\linewidth]{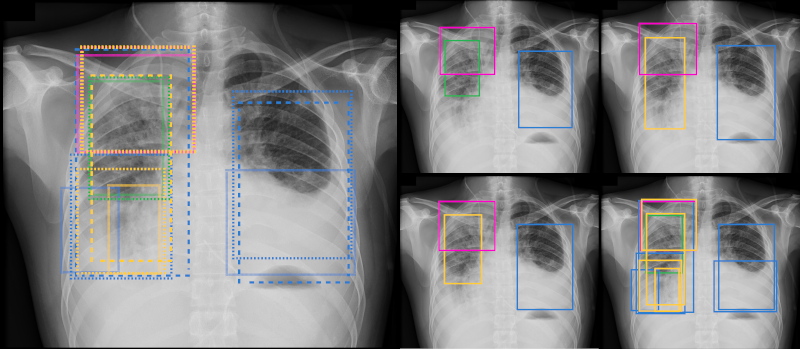}
\end{center}
\vspace{-0.75cm}
   \caption{Comparison between different ground truth aggregation methods, exemplary on the VinDr-CXR dataset \cite{nguyen2022vindr}. Left: the original image with the repeated labels indicated by the different line types. Right: the four smaller images from top left to bottom right are, MJV+$\cap$, LAEM+$\mu$, LAEM+$\cup$ and WBF.}
\label{fig:long}
\label{fig:onecol}
\end{figure}

Data-driven machine learning systems are expected to operate effectively even under "difficult" and unforeseen circumstances. Consider safety-relevant domains such as autonomous driving, medical diagnosis, or structural health monitoring, where system failure sets lives at risk. Robust systems $-$ those capable of reliable operation in unseen situations $-$ may encounter several challenges, including domain shifts \cite{robust_learning_domain_shift,robust_learning_domain_shift_2,robust_learning_domain_shift_3,robust_learning_domain_shift_4}, adversarial attacks \cite{robust_learning_adversarial,robust_learning_adversarial_2}, degrading image quality \cite{robust_learning_image_quality_datasets,robust_learning_image_quality_marine_waste,robust_learning_image_quality_drones} and noisy or uncertain labels \cite{vindr_dataset,robust_learning_jiou,robust_learning_noise_tolerant,robust_learning_pseudo_labels}.  Past studies \cite{survey_noisy_labels} indicate that noisy labels can cause more harm than the three aforementioned sources of input noise. Given this context, our study concentrates on addressing the issue of noisy labels, specifically within noisy test data. Without a unique ground truth, evaluation is unattainable. Therefore, to enhance robustness against label noise, it will be pivotal to first devise methods tailored towards annotation aggregation, which lays the groundwork for potential future integration with multi-annotator learning methods.

The creation of annotated data for supervised learning is a costly endeavor, particularly in cases where experts such as medical professionals or domain experts are needed to annotate the data. To mitigate this issue, crowd-sourcing has emerged as a cost-effective means of generating large datasets, albeit with the disadvantage of potentially lower quality annotations that may contain label noise \cite{ml_crowdsourcing,distilling_severe_label_noise,noise_types}. Although the reduced costs of crowd-sourced annotations often justifies their use, deep neural networks have the capacity to memorize noisy labels as special cases, leading to a declining performance and overfitting towards the noisy labeled data \cite{distilling_severe_label_noise}. Notably, even expert annotated data is susceptible to label noise, given the difficulty of the data to annotate. A survey by Song \etal \cite{survey_noisy_labels} revealed that the number of corrupt labels in real-world datasets ranges between 8.0\% to 38.5\%. The authors demonstrate that reducing label noise and creating cleaned data can improve the accuracy of models. To address the issue of noisy labels, an approach known as ``repeated-labeling'' has been proposed. Repeated-labeling means to obtain annotations from multiple annotators/coders for the same data entry, such as an image. More specifically: For a set of images $\{x_i\}^N_{i=1}$ multiple annotators create noisy labels $\{\Tilde{y}_i^{r}\}^{r=1,\dots,R}_{i=1,\dots,N}$, with $\Tilde{y}_i^{r}$ being the label assigned from annotator $r$ to image $x_i$, but without a ground truth label $\{y_i\}_{i=1,\dots,N}$ \cite{tanno2019learning}. 

\begin{figure*}[ht]
\begin{center}
    \includegraphics[width=1.0\textwidth]{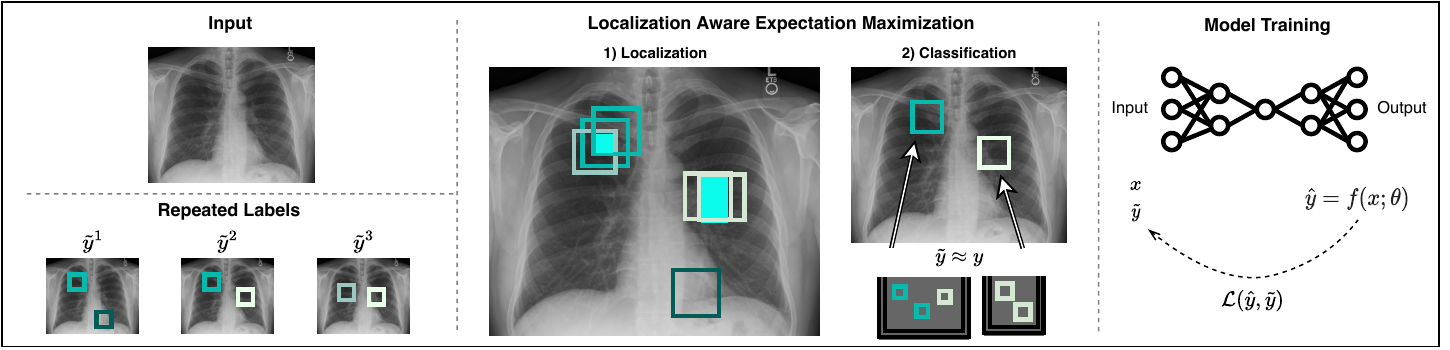}
    \vspace{-0.75cm}
    \caption{Left: Original input image featuring three separate annotations by distinct annotators. Center: Application of the LAEM aggregation method to the three annotations, yielding an approximate ground truth. Right: Aggregated ground truth utilized during the training process.}
    \label{fig:processing_graphic}
\end{center}
\end{figure*}

Methods for mitigating the negative effect of label noise via repeated labeling can be divided into two categories \cite{vindr_dataset,tanno2019learning}: (a) \emph{two-stage} approaches \cite{glad,dawid_skene} and (b) \emph{one-stage} or \emph{simultaneous} approaches \cite{mbem,wdn,gao_data_dependent}. Two-stage approaches aim to approximate the ground truth prior to training, a process known as ground truth estimation or ground truth inference \cite{crowdsourcing_GTI}, as depicted in Figure \ref{fig:processing_graphic}; a straightforward approach is to compute a majority vote. Following label aggregation, the model is trained in a regular fashion. Two-stage approaches offer the benefit of being compatible with commonly used model architectures. On the other hand, simultaneous approaches attempt to integrate repeated labels directly into the training process. In any case, the primary objective of both strategies is to achieve robust and accurate results by leveraging the repeated labeled data to the fullest extent possible. Doing so is crucial to justify the additional annotation efforts. Lastly, to enable the use of established performance metrics, such as those employed in the COCO object detection dataset (mAP) \cite{coco}, a ground truth estimation step is essential for the validation and test sets. While simultaneous approaches can more effectively utilize repeated labels, they are not intended to execute the necessary aggregation step required to generate the unique ground truth estimate \cite{tanno2019learning}. Consequently, reliable approximation methods are indispensable for evaluation purposes.

Object detection and instance segmentation require both localization and classification, which means that existing methods for repeated labels that are used for classification tasks such as image classification or named entity recognition are not applicable \cite{dl_f_crowds}. That is, the available selection of ground truth inference methods is limited. Furthermore, the creation of bounding box or polygonal annotations is expensive \cite{robust_learning_noise_tolerant} and reduces the number of datasets with repeated labels available for evaluating ground truth inference methods \cite{own_layout_paper,nguyen2022vindr}. However, we deliberately avoid using synthetic data and focus on real datasets. Our contributions are as follows:
\begin{enumerate}
\item
We propose a localization algorithm that enables the use of existing ground truth estimation methods such as majority voting or expectation maximization for instance-based recognition tasks and evaluate it extensively with existing methods \cite{wbf,vindr_dataset}.
\item
We introduce a comparative analysis of ground truth inference methods that highlights their properties and limits.
\item
We conduct ablation studies to analyze the costs associated with creating repeated annotations, and what to do when the amount of available annotated data is limited.
\item
We introduce an extension for the TexBiG dataset \cite{own_layout_paper} in the form of a test subset, wherein each of the 200 test images has been annotated by five expert annotators. Utilizing our aggregation method, we establish a unique approximation of the ground truth, which will serve as the unknown reference standard on an evaluation server. This approach allows the TexBiG dataset to be used for evaluation of robust learning methods addressing the challenge of noisy labels.
\end{enumerate}

Once released, the link to the evaluation server will be posted on the GitHub repository where the code is hosted: \url{https://github.com/Madave94/gtiod}.

%%%%%%%%%%%%%%%%%%%%%%%%%%%%%%%%%%%%%%%%%%%%%
\section{Related Work}

To approximate the ground truth, estimation methods make assumptions about the data and task properties as well as the annotation process. Majority Voting (MJV) \cite{dl_noisy_labels_medical,mjv_1,mjv_2} assumes correct labels  for the majority of training samples and aggregates the labels accordingly:
\begin{equation}
    \Tilde{y_i} = 
    \begin{cases}
        \text{$1$ if $(1/R) \sum^R_r= y_{i}^r > 0.5$}  \\
        \text{$0$ if $(1/R) \sum^R_r= y_{i}^r < 0.5$} 
    \end{cases}
    \label{mjv_formula}
\end{equation}
In case of a tie, the label is chosen randomly between the tied ones or selected by a super-annotator.  On data with high inter-annotator agreement, majority voting can be a straightforward approach to obtain ground truth estimates reasonable quality.

Numerous methods for inferring ground truth rely on the Expectation-Maxi\-mi\-za\-tion (EM) approach, first introduced by Dawid and Skene \cite{dawid_skene}. This approach estimates annotator confidence and integrates it into a weighted voting procedure for determining the true label. By considering annotator performance, these methods address the limitations of majority voting, thereby avoiding potential outliers. One notable advancement in this area is the GLAD \cite{glad} method, which not only attempts to identify the most probable class but also assesses image difficulty, additional to the annotator confidence. It should be noted, however, that this approach is limited to binary classification tasks \cite{fast_dawid_skene}.

In addition to classification tasks, pixel-wise classification (semantic segmentation) also has existing ground truth inference methods, such as STAPLE \cite{staple_method}, SIMPLE \cite{simple_method}, and COLLATE \cite{collate_method}. Recent developments in this field have led to approaches that incorporate data difficulty into the estimation process, as seen in a newly developed simultaneous method \cite{gao_data_dependent}. Although there are numerous variations of ground truth estimation methods for classification and segmentation tasks, this discussion will focus on methods applicable to object detection and instance segmentation, rather than diving deeper into this area.

For instance-based recognition tasks like object detection and instance segmentation, there is an additional issue to consider -- the localization step. During training, methods consisting of a combination of thresholds and non-maximum suppression are used to solve the localization problem and then focus on classification accuracy. While this may work during training, repeated labeling is likely to have more than just a prediction and ground truth pair to match, since multiple annotators might have created multiple labels. Hence, existing methods are not applicable. An existing approach to aggregate annotations for object detection is called Weighted Boxes Fusion (WBF) \cite{wbf,vindr_dataset}, which was used for the VinDr-CXR dataset \cite{nguyen2022vindr}. WBF focuses on the weighted aggregation within each class, ignoring inter-class disagreements and also not discarding any annotations even with low agreement. This is beneficial in cases where missing a possible case is far more severe then finding too many, such as a task that requires high recall. Apart from this single existing instance-based recognition approach, we are not aware of any other aggregation methods for object detection or instance segmentation.

%%%%%%%%%%%%%%%%%%%%%%%%%%%%%%%%%%%%%%%%%%%%%
\section{Method}

In the following section we introduce a novel adaptation of the EM algorithm, \textit{localization-aware expectation maximization} (LAEM), for instance-based recognition tasks. The same localization algorithm can also be used with majority voting, which therefore functions as a baseline.  Additionally, we expand the existing weighted boxes fusion technique to encompass weighted mask fusion, which enables its use in instance segmentation and facilitates benchmarking on a broader range of datasets. As extending weighted boxes fusion is not our core contribution it can be found in \nameref{appx_1}.

\subsection{Localization-Aware Expectation-Maximization}
Our novel approach adds a localization stage to existing methods like majority voting and expectation maximization, enabling the use of these established methods for instance-based recognition tasks. Thus, the proposed label aggregation process consists of two stages: (1) \textit{localization-stage} and (2) \textit{classification-stage}. Assuming that $R$ annotators have created noisy instance-based labels $\Tilde{y}_{ij}^r$ for image $x_i$.  Subscript $j=0,...,M_r$ refers to the single instances annotated by annotator $r$ for image $x_i$. $M_r$ can be zero if no instances were labeled by $r$. Each instance contains a class $c \in C$ denoted $\Tilde{y}_{ijc}^r$. Furthermore, $\Tilde{y}_{ijb}^r$ refers to the respective bounding box and $\Tilde{y}_{ijs}^r$ to the optional pixel-wise segmentation mask. 

\begin{algorithm*}[ht]
\scriptsize
     \caption{Outline of the localization algorithm used for LAEM}\label{alg:cap}
     \begin{algorithmic}
         \Require

         \State $X = \{x_i\}_{i=1, \dots, N}$     \Comment{Set of images}
         \State $\tilde{Y} = \{\tilde{Y}_i\}_{i=1, \dots, N}$     
\Comment{Set of noisy labels per image}
         \State $S = \{S_i\}_{i=1, \dots, N}$     \Comment{Set of annotators 
per image}
         \State $\theta$     \Comment{IoU threshold}
         \For{$i\in X$}     \Comment{Loop over images}

             \State $\tilde{Y}_i = \{\tilde{y}^{1}_{i1}, 
\tilde{y}^{1}_{i2}, \dots, \tilde{y}^{1}_{iM_1}, \tilde{y}^{2}_{i1}, 
\tilde{y}^{2}_{i2}, \dots, \tilde{y}^{2}_{iM_2}, \dots, 
\tilde{y}^{R}_{i1}, \tilde{y}^{R}_{i2}, \dots, \tilde{y}^{R}_{iM_R}\}$
             \State $\tilde{Y}_i^{\text{LAEM}} = \emptyset$

             \State $Q$ = $\{U_k | U_k \in \mathcal{P}(S_i) \wedge |U_{k}| 
\geq |U_{k+1}| \wedge \lceil|S_i| / 2\rceil \leq |U_k|\}$
             \Comment{Ordered set of annotator combinations}

             \For{$U \in Q$}     \Comment{Loop over annotator combinations}
                 \State L = $\{ \tilde{Y}_{i:}^{k_1} \times \cdots 
\times \tilde{Y}_{i:}^{k_n} | k_1, \dots, k_n \in U \wedge n = |U| \}$
                 \Comment{Possible combinations of labels}
                 %\State L = $\times_{i=1}^{} c$
                 \State $F = \{u_k | u_k \in L \wedge \theta \leq 
IoU(u_k) \wedge IoU(u_k) \geq IoU(u_{k+1})\}$
                 \Comment{Filtered and ordered L}
                 \For{$k \in N$}     \Comment{Loop over label combinations}
                     \State $K = \{k_1, k_2, \dots, k_n\}$
                     \If{$ K \cap \tilde{Y}_{i} = \emptyset$}     
\Comment{Check for label availability}
                     \State $\tilde{Y}_i = \tilde{Y}_i \setminus K$  
\Comment{Remove labels from available labels}
                     \State $\tilde{Y}_i^{\text{LAEM}} = 
\tilde{Y}_i^{\text{LAEM}} \cup \emph{aggregate}(K)$
     \Comment{Add aggregated label to accepted labels}
                     \EndIf
                 \EndFor
             \EndFor
         \EndFor

     \end{algorithmic}
     \label{algo}
\end{algorithm*}
Algorithm\,\ref{algo} outlines the LAEM approach. The algorithm requires image set $X$, a set of noisy labels $\tilde{Y}$, a set of annotators $S$, and a threshold $\theta$. Looping over the images of the dataset, the power set $\mathcal{P(S)}$ over the annotators is computed. Subsets containing less than half the number of annotators are removed and a descending order is enforced onto the set. It subsequently iterates through the remaining ordered subsets of annotators and computes the Cartesian product between the respective annotators. Each tuple is then ordered and filtered according to threshold $\theta$ based on the intersection over union in its generalized form:
\begin{equation}
    IoU = \dfrac{\bigcap_{r=1}^R \Tilde{y}_{ijb}^r} { \bigcup_{r=1}^R \Tilde{y}_{ijb}^r }
\end{equation}
The remaining set of tuples ordered by descending IoU forms the set of candidate solutions $F$. In case all labels from a candidate tuple are still available, they are aggregated according to an aggregation function and  added to the inferred solutions $\tilde{Y}_i^{\text{LAEM}}$. The aggregation function comprises two steps: (1) all classes contained in the original tuple $\tilde{y}^{r}_{ijc}$ are appended to a list serving as input for expectation maximization or majority voting and (2) the areas of the different candidates $\tilde{y}^{r}_{ijb}$ are combined according to the union, intersection, or average area of all boxes involved. The average operation is based on the WBF algorithm \cite{wbf} with uniform weights. If available, the same procedure (for details cf.\ \nameref{appx_1}) is applied to the segmentation masks $\tilde{y}^{r}_{ijs}$. This concludes the localization stage. In the subsequent classification-stage existing ground truth inference methods can be applied such as majority voting or expectation maximization \cite{dawid_skene}.

\subsection{Algorithmic Design Choices}
\label{algo_design}

Our algorithm is designed in a divide-and-conquer manner. Firstly, we prioritize localization, effectively reducing the problem to a classification task for each matched instance after localization. This strategy consequently facilitates the application of established methods for ground truth inference developed in other fields. We always prefer a localization match with more annotators to maximize consensus. If a localization match involving all available annotators cannot be found given the threshold value $\theta$, we ensure successive reduction to potentially prefer the next largest number of annotators. This approach first guarantees localization quality, and only upon establishing matched areas based on their localizations do we aggregate the classes. The algorithm is parameterized by the threshold value $\theta$, which can be adjusted to enforce stricter localization quality and also control the order in which instances are matched. Though this heuristic solution may not provide an optimal outcome for larger problem sizes (e.g., numerous instances on a single image), when an image exhibits high agreement among annotators, a consensus area can be aggregated, and the class of this area can be unambiguously determined.

One advantage of the Expectation-Maximization (EM) approach is that assignment is unambiguous. The confidence calculated during the EM algorithm serves as a tie-breaker, a benefit not present with Majority Voting (MJV). Furthermore, fitting the EM algorithm is efficient; following localization matching, no further areas are calculated, and only the solutions $\tilde{Y}_i^{\text{LAEM}}$ are considered along with their classes.

While localization fusion functions, such as union or intersection, are available and applicable for training data, the intended use for test data within the context of LAEM (Localization-Aware Expectation-Maximization) primarily involves the averaging fusion function. This approach enables a balanced aggregation of areas across different annotators. Additionally, this method is also utilized to aggregate test data as required for the TexBiG dataset \cite{own_layout_paper}.

%%%%%%%%%%%%%%%%%%%%%%%%%%%%%%%%%%%%%%%%%%%%%
\subsection{Comparative Analysis}

\begin{table*}[]
%\scriptsize 
\fontsize{7}{10}\selectfont % no effect for fractal font sizes (i.e. 7.5pt); standard linespace would be 9, here increased to 10
\begin{center}
\def\colwidth{18.8ex}
\begin{tabular}{l|>{\centering\arraybackslash}p{25ex}>{\centering\arraybackslash}p{\colwidth}>{\centering\arraybackslash}p{\colwidth}>{\centering\arraybackslash}p{\colwidth}}
\hline
\toprule
Methods & LAEM & MJV & WBF & WBF+EM \\ \hline
Assignment & \begin{tabular}[c]{@{}l@{}}1) Localization\\ 2) Classification\end{tabular} & \begin{tabular}[c]{@{}l@{}}1) Localization\\ 2) Classification\end{tabular} & \begin{tabular}[c]{@{}c@{}}Localization\\ only\end{tabular} & \begin{tabular}[c]{@{}c@{}}Localization\\ only\end{tabular}  \\ \hline
Low agreement & Discard annotation & Discard annotation & Keep annotation & Keep annotation \\ \hline
Edge cases & Use confidence & Randomized & \multicolumn{1}{c}{$-$} & \multicolumn{1}{c}{$-$} \\ \hline
Localization fusion & \begin{tabular}[c]{@{}c@{}}Union / average /\\ intersection\end{tabular} & \begin{tabular}[c]{@{}c@{}}Union / average /\\ intersection\end{tabular} & Averaging & \begin{tabular}[c]{@{}c@{}}Weighted\\ averaging\end{tabular} \\ \hline
Annotator confidence & \multicolumn{1}{c}{\checkmark} & \multicolumn{1}{c}{$\times$} & \multicolumn{1}{c}{$\times$} & \multicolumn{1}{c}{\checkmark} \\ \hline
Handling missing data & \multicolumn{1}{c}{\checkmark} & \multicolumn{1}{c}{\checkmark} & \multicolumn{1}{c}{\checkmark} & \multicolumn{1}{c}{$\checkmark$} \\ \hline
Dataset characteristic & Balanced & Precision oriented & Recall oriented & Recall oriented \\ \hline
Data dependence & \multicolumn{1}{c}{$\times$} & \multicolumn{1}{c}{$\times$} & \multicolumn{1}{c}{$\times$} & \multicolumn{1}{c}{$\times$} \\ 
\bottomrule
\end{tabular}
\end{center}
\caption{Comparison table for the characteristics and properties of the different ground truth inference methods. MJV and LAEM both use the novel localization algorithm.}
\label{taxonomy}
\end{table*}

In Table\,\ref{taxonomy}, we present a comparative analysis of the four available ground truth inference methods for instance-based recognition tasks, distinguished by their respective characteristics and properties. Each method is described based on eight distinct features relevant to ground truth estimation methods. A noteworthy difference between LAEM and MJV, as compared to WBF and its adaptation (detailed in \nameref{appx_2}), is the handling of instances that lack consensus among annotators. Figure \ref{fig:long} and \nameref{appx_3} illustrates the aggregation processes of MJV, LAEM, and WBF on a few specific images, serving as practical examples. This illustrations reveal that MJV and LAEM tend to find consensus instances, resulting in a final image that appears as if a single annotator has labelled the image. In contrast, the WBF image is relatively cluttered with overlapping instances. This discrepancy arises because WBF merges areas of the same class that significantly overlap but does not discard any annotation. This is also the case, for instances where two annotators found the same instance area but disagreed on the class, resulting in more instances overall. Although this property might be beneficial for a high-recall scenario $-$ where missing an instance is more detrimental than detecting multiple false positives $-$ it is not ideal for many applications.
It's important to note that none of the current methods incorporate data dependence, a feature described in several state-of-the-art ground truth estimation methods for semantic segmentation \cite{simple_method,staple_method,collate_method}.

\section{Experimental Results}

In our preliminary experiment, we scrutinized the influence of annotation budget size by exploring scenarios in which repeated labels might be preferred over single labels. This ablation study was designed to determine the optimal use of a restricted annotation budget, i.e., whether it is more beneficial to train with a larger volume of noisy labels or a reduced set of refined labels. This experimental analysis was conducted using two separate datasets.

In our subsequent investigation, we assessed the effect of annotator selection on model performance by deliberately excluding certain annotators from the labeling process. This enabled us to emulate the potential impact of annotator selection on model performance and to probe the influence of proficient and suboptimal annotators on model output.

Our final experiment, which is detailed in \nameref{appx_2}, was actually conducted first since it influenced the choice of the aggregation method used for the training data. However, this experiment is not the main focus of this publication. 

\subsection{Set-Up}

To the best of our knowledge, there are only two datasets available that contain repeated labels for object detection and instance segmentation, respectively: the VinDr-CXR \cite{nguyen2022vindr,vindr_dataset} dataset and the TexBiG \cite{own_layout_paper} dataset. We focus solely on these two datasets and do not make use of any synthetic data.

\textbf{VinDr-CXR dataset.} This dataset comprises 15,000 training and 3,000 test chest X-ray images, each of which was annotated by three annotators and five annotators, respectively. With a total of 36,096 instances in the training dataset, the dataset can be considered sparsely annotated, with in average $0.8$ instances per image\footnote{Computed by dividing the number of instances by the product of the number of images and the number of annotators.}. The dataset consists of 14 different classes that were annotated by 17 radiologists \cite{vinbigdata-kaggle}. Using the agreement evaluation method presented in \cite{own_layout_paper} describing the data quality, the K-$\alpha$ (Krippendorff's alpha) is $0.79$. However, since only $29.3\%$ of the images in the dataset contains any annotations at all, another K-$\alpha$ was calculated for this reduced subset, resulting in a K-$\alpha$ value of $0.29$. This indicates that while annotators largely agree in cases where no anomaly is present, there is significant disagreement in cases that contain instances.

\textbf{TexBiG dataset.} Recently published \cite{own_layout_paper}, the TexBiG provides labels for document layout analysis, similar to the PubLayNet \cite{publaynet} or DocBank \cite{docbank} datasets. It covers 19 classes for complex document layouts in historical documents during a specific time period, and in the version used here, the training data contains 44,121 instances, the validation data 8,251 instances and the test data 6,678 instances. While the total number of instances is larger as in the VinDr-CXR dataset, there are only 2,457 images in total, 1,922 in the training set, 335 in the validation set and 200 in the test set. Due to the iterative creation process of the dataset, the number of repeated labels is different depending on the sample. An agreement value was used per image to evaluate which samples were to be annotated again. For each image, two annotators were assigned, and in case the agreement value was low after the first iteration, an additional annotator was added to that specific sample. This was done until a maximum of four annotators per sample. In the combined validation and training set, 34 images were annotated by 4 annotators, 336 by at least 3 annotators (including the 34 from before), and 2,257 by at least 2 annotators.  We created an additional test set with 5 annotators for 200 newly selected images from the same domain, in accordance with the guideline provided by the authors \cite{own_layout_paper}. We plan to publish this test-set for benchmarking purposes on an evaluation server. The TexBiG dataset is more densely annotated, with $10.7$ instances per image, which is $13$ time more than the VinDr-CXR dataset. Furthermore, the K-$\alpha$ for the TexBiG training dataset is higher with $0.93$.

Comparing the two datasets we find that they represent two opposing marginal cases with one dataset having high-agreement and dense annotations, while the other one has a low-agreement and sparse annotations. However, a more balanced dataset is missing.

\textbf{Architecture choice.} Regarding the architecture choice, we aimed to find a well-performing and stable choice, rather than aiming for state-of-the-art results since we wanted to focus on comparing the ground truth inference methods and ablation studies on different tasks. For the VinDr-CXR dataset, we tested various architectures including different anchor-based two-stage detectors like Faster R-CNN \cite{faster_rcnn}, Cascade R-CNN \cite{cascade_rcnn} and Double Head R-CNN \cite{double_head}, and additionally, the transformer-based Detection Transformer (DETR) \cite{detr}. After thorough investigation, we found that the Double Head R-CNN performs stably and with reasonable results. Therefore, we selected this architecture for our experiments. On the TexBiG dataset, we tried several instance segmentation models like Mask R-CNN \cite{mask_rcnn}, Cascade Mask R-CNN \cite{cascade_rcnn} and DetectorRS \cite{detecto_rs}, as well as the Mask2Former \cite{mask2former} as a transformer-based architecture. In this case, DetectoRS yielded the most stable performance, and we continued our experiments with this model. We extended MMDetection \cite{mmdetection} with our implementation, and the code is available on GitHub under \url{https://github.com/Madave94/gtiod}.

 \subsection{Annotation Budget Ablation}
 \label{contingent_study}

 When working on a deep learning project, data is often a limiting factor, and resources must be carefully allocated to create task-specific data. Even when there are ample resources, maximizing the value of those resources is important. In the context of human-annotated data, this concept is referred to as the ``annotation budget'', which represents the available number of images or instances that can be labeled by a pool of annotators within their available time. The question then becomes, ``How can a limited annotation budget be best utilized?'' One approach is to prioritize annotating as many different images as possible to cover a broad range of cases within the application domain. However, this approach comes with the risk of introducing more noisy labels due to the inherent variability in annotator performance. Alternatively, creating repeated labels may be more beneficial to improve the quality of the annotations. Ultimately, the decision between prioritizing \textit{quantity versus quality} of labels must be carefully weighed and considered in the context of the project goals and available resources.

 \begin{table}[]
 \centering
 \begin{minipage}{.5\linewidth}
  \centering
\begin{tabular}{>{\raggedleft\arraybackslash}p{8ex}|rl|cc|cc}
\toprule
\multirow{2}{*}{Split}                                                     & \multicolumn{2}{l|}{Budget} & \multicolumn{2}{l|}{Averaged} & \multicolumn{2}{l}{Maximum} \\
                                                                           & \multicolumn{1}{c}{rel.}            & \multicolumn{1}{c|}{abs.}          & AP             & AP$^{bb}$         & AP           & AP$^{bb}$         \\ \hline
1922$\times$2                                                                   & 100\%           & 3844          & 41.9           & 47.5         & 43.3         & 48.7         \\ \hline
\begin{tabular}[c]{@{}r@{}}966$\times$2\\ 966$\times$1\end{tabular}                   & 75\%            & 2883          & 42.4          & 47.9         & 42.8         & 48.4         \\ \hline
1922$\times$1                                                                   & 50\%            & 1922          & \textbf{42.7}           & \textbf{48.3}         & 43.4         & \textbf{48.8}         \\ \hline
\begin{tabular}[c]{@{}r@{}}641$\times$2\\ 640$\times$1\end{tabular}                  & 50\%            & 1922          & 42.4           & 47.7         & \textbf{43.9}         & \textbf{48.8}         \\ \hline
966$\times$2                                                                    & 50\%            & 1922          & 41.1           & 46.2         & 42.8         & 47.8         \\ \hline
\begin{tabular}[c]{@{}r@{}}30$\times$4\\ 243$\times$3\\ 1073$\times$1\end{tabular}        & 50\%            & 1922          & 41.9           & 47.3         & 43.0           & 48.2         \\ \hline
\begin{tabular}[c]{@{}r@{}}30$\times$4\\ 243$\times$3\\ 536$\times$2\\ 1$\times$1\end{tabular} & 50\%            & 1922          & 39.7           & 45.0         & 41.0           & 46.5 \\
\bottomrule
\end{tabular}
\caption{Ablation study on the TexBiG dataset using a limited annotation budget. The results are in $mAP@[.5:.95]$, show that multi annotator learning methods are required to justify repeated labels. However, even without multi annotator methods the performance loss using repeated annotations is marginal.}
\label{texbig_ann_contingent}
\end{minipage}%
\hspace{.03\linewidth}
\begin{minipage}{.45\linewidth}
 \centering
\begin{tabular}{r|cc|c|c}
\toprule
\multicolumn{1}{c|}{\multirow{2}{*}{Split}}                                         & \multicolumn{2}{c|}{Budget} & \multirow{2}{*}{\begin{tabular}[c]{@{}c@{}}Avg.\\ AP\end{tabular}} & \multirow{2}{*}{\begin{tabular}[c]{@{}c@{}}Max.\\ AP\end{tabular}} \\
                                                               & rel.           & abs.           &                                                                    &                                                                    \\ \hline
15,000$\times$2                                                     & 66.6\%         & 30,000         & 14.8                                                               & 15.0                                                               \\ \hline
10,000$\times$3                                                     & 66.6\%         & 30,000         & 14.7                                                               & 14.9                                                               \\ \hline
15,000$\times$1                                                     & 33.3\%         & 15,000         & 13.4                                                               & 13.9                                                               \\ \hline
\begin{tabular}[c]{@{}r@{}}10,000$\times$1\\ 2,500$\times$2\end{tabular} & 33.3\%         & 15,000         & \textbf{13.6}                                                               & 14.1                                                               \\ \hline
7,500$\times$2                                                      & 33.3\%         & 15,000         & 13.5                                                               & 13.8                                                               \\ \hline
\begin{tabular}[c]{@{}r@{}}3,000$\times$3\\ 3,000$\times$2\end{tabular}  & 33.3\%         & 15,000         & \textbf{13.6}                                                               & \textbf{14.3}                                                               \\ \hline
5,000$\times$3                                                      & 33.3\%         & 15,000         & 13.4                                                               & 14.0      \\
\bottomrule 
\end{tabular}
\caption{Ablation study on the VinDr-CXR dataset using a limited annotation budget. The results are in $mAP_{40}$ as provided by the leaderboard.}
\label{vindrcxr_ann_contingent}
\end{minipage} 
\end{table}

In the two ablation studies presented in Table \ref{texbig_ann_contingent} and \ref{vindrcxr_ann_contingent}, we compare the performance of different annotation budgets, which refer to the available number of images or instances that can be labeled with the pool of annotators and their available time. The splits used in the studies represent three different cases: (1) Only single annotated labels are available, which are more prone to label noise. (2) A mix of repeated labels and single annotated labels is available. Multiple splits may have this property. (3) Maximum label repetition, where no or very few single annotated labels are available, resulting in significantly less training data. To reduce randomization effects, we create five different random versions for each split and compute their mean and maximum results.

Our results show that the TexBiG dataset quickly reached a data saturation point, suggesting potential benefits from employing multi-annotator learning methods to better utilize repeated labels. Conversely, the VinDr-CXR dataset showed improved performance with higher budgets, indicating that more data helps performance in scenarios with noisy, low-agreement labels.

Both datasets demonstrate that moderate inclusion of repeated labels does not adversely impact performance, with mixed splits achieving peak results at their lowest budgets. These findings highlight the value of repeated annotations, which not only increase label reliability, but also allow for efficient use of multi-annotator learning methods. Remarkably, the opportunity costs for creating such repeated labels seem negligible.

Our findings suggest that higher fragmentation in annotator splits could lead to reduced performance, possibly due to enhanced intracoder consistency. Moreover, the influence of split distribution appears prominent only when the annotation budget is limited. Identifying a systematic relationship between split distribution and performance, thereby suggesting optimal splits before the annotation process, could be a promising future research direction.

The overall takeaway is that multiple annotations may not always yield significant advantages, yet in scenarios with a constrained annotation budget, they could prove beneficial. Determining which cases fall into each category remains an open challenge.

\subsection{Leave-One-Out Annotator Selection}
\label{leave_out_study}

Table \ref{leave_one_out} displays the results of a final experiment conducted on the TexBiG dataset. To create four groups of annotators, each group consisting of one to three individuals, annotations were distributed unevenly among them, resulting in groups of different sizes. Subsequently, each group was left out of the training process, while the remaining three groups were used to train the model. This approach led to a smaller training set. Surprisingly, the experiment showed that when the largest group, denoted as \textbf{B}, was excluded, leaving only $61.6\%$ of the annotations available, the model's performance reached its peak. This outcome underscores the importance of selecting precise annotators in the training process, since less precise ones may introduce noisy labels that can hinder performance. However, it is challenging to identify precise annotators before the annotation process, as there is no data available to determine their level of precision.

\begin{table}[]
\centering
\setlength{\tabcolsep}{6pt}
\begin{tabular}{l|rr|rr|cc}
\toprule
\multirow{2}{*}{\begin{tabular}[c]{@{}c@{}}Left out \\ group\end{tabular}} & 
\multicolumn{2}{c|}{\begin{tabular}[c]{@{}c@{}}Left out \\ images\end{tabular}} & 
\multicolumn{2}{c|}{\begin{tabular}[c]{@{}c@{}}Left out \\ annotations\end{tabular}} & 
\multicolumn{2}{c}{\begin{tabular}[c]{@{}c@{}}Perfor- \\ mance\end{tabular}}                   
\\
& 
\multicolumn{1}{c}{rel.} & 
\multicolumn{1}{c|}{abs.} & 
\multicolumn{1}{c}{rel.} & 
\multicolumn{1}{c|}{abs.} & 
\multicolumn{1}{c}{AP} & 
\multicolumn{1}{c}{AP$^{bb}$} \\ \hline
Group A & 25.1\% & 1,040 & 26.8\% & 11,810 & 42.4 & 47.1 \\
Group B & 29.5\% & 1,225 & 38.4\% & \multicolumn{1}{r|}{16,932} & \textbf{44.1} & \textbf{49.8} \\
Group C & 25.7\% & \multicolumn{1}{r|}{1,067} & 18.2\%  & \multicolumn{1}{r|}{8,017} & 42.6 & 48.1 \\
Group D & 19.7\% & \multicolumn{1}{r|}{815} & 16.7\% & \multicolumn{1}{r|}{7,362} & 43.1 & 48.3 \\
\bottomrule
\end{tabular}
\caption{Choosing the right annotator? If annotators are not in the group what would happen to the results? Splits are unequal, due to the annotation distribution.}
\label{leave_one_out}
\end{table}

%%%%%%%%%%%%%%%%%%%%%%%%%%%%%%%%%%%%%%%%%%%%%
\section{Conclusion}

%This paper presents a novel localization algorithm, a unique feature that allows the transformation of combined localization and classification tasks into classification-only problems. This capability enables the application of techniques such as Expectation-Maximization (EM) or Majority Voting (MJV), originally developed for classification tasks, in a new and promising context.

Our results indicate the potential benefits of repeated labels which seem to be contingent on several factors. The identified key factors are the balance between (1) the complexity or variation in the dataset and its corresponding task difficulty, (2) the variability in annotation depending on inter-annotator consistency and annotator proficiency, and (3) the constraints of the annotation budget. This interaction suggests the existence of an `optimal range' for image annotation strategy. For instance, datasets with high variance and low annotator consistency may benefit from multiple annotations per image, while in cases with low image variation and high annotator consistency, many images annotated once might suffice. This balancing act between data and annotation variation could guide decisions when choosing between single or multiple annotators per image, given a fixed annotation budget.

However, the utility of repeated labels is substantially hampered due to the lack of multi-annotator-learning approaches for object detection and instance segmentation. Thus, future work should concentrate on developing methods that bridge this gap between these areas and other computer vision domains like image classification or semantic segmentation.

Lastly, a significant challenge remains regarding the availability of suitable datasets. With limited datasets in the domain and disparities among them, our findings' generalizability remains constrained to the two domains covered in this study. A larger dataset with repeated labels and balanced agreement would be valuable for future research. Synthetic data could be beneficial but pose the risk that models trained on these data may only learn the distribution used to randomly create repeated labels from the original annotations. Thus, creating a suitable dataset remains a formidable task.

\subsubsection{Acknowledgments.}
This work was supported by the Thuringian Ministry for Economy, Science and Digital Society / Thüringer Aufbaubank (TMWWDG / TAB).

\bibliographystyle{splncs04}
\bibliography{090-main}

\begin{thebibliography}{10}
\providecommand{\url}[1]{\texttt{#1}}
\providecommand{\urlprefix}{URL }
\providecommand{\doi}[1]{https://doi.org/#1}

\bibitem{collate_method}
Asman, A.J., Landman, B.A.: Robust statistical label fusion through consensus
  level, labeler accuracy, and truth estimation (collate). IEEE transactions on
  medical imaging  \textbf{30}(10),  1779--1794 (2011)

\bibitem{cascade_rcnn}
Cai, Z., Vasconcelos, N.: Cascade r-cnn: High quality object detection and
  instance segmentation. IEEE Transactions on Pattern Analysis and Machine
  Intelligence p. 1–1 (2019). \doi{10.1109/tpami.2019.2956516},
  \url{http://dx.doi.org/10.1109/tpami.2019.2956516}

\bibitem{detr}
Carion, N., Massa, F., Synnaeve, G., Usunier, N., Kirillov, A., Zagoruyko, S.:
  End-to-end object detection with transformers. In: Computer Vision--ECCV
  2020: 16th European Conference, Glasgow, UK, August 23--28, 2020,
  Proceedings, Part I 16. pp. 213--229. Springer (2020)

\bibitem{mmdetection}
Chen, K., Wang, J., Pang, J., Cao, Y., Xiong, Y., Li, X., Sun, S., Feng, W.,
  Liu, Z., Xu, J., Zhang, Z., Cheng, D., Zhu, C., Cheng, T., Zhao, Q., Li, B.,
  Lu, X., Zhu, R., Wu, Y., Dai, J., Wang, J., Shi, J., Ouyang, W., Loy, C.C.,
  Lin, D.: {MMDetection}: Open mmlab detection toolbox and benchmark. arXiv
  preprint arXiv:1906.07155  (2019)

\bibitem{robust_learning_domain_shift_4}
Chen, Y., Li, W., Sakaridis, C., Dai, D., Van~Gool, L.: Domain adaptive faster
  r-cnn for object detection in the wild. In: Proceedings of the IEEE
  conference on computer vision and pattern recognition. pp. 3339--3348 (2018)

\bibitem{mask2former}
Cheng, B., Misra, I., Schwing, A.G., Kirillov, A., Girdhar, R.:
  Masked-attention mask transformer for universal image segmentation. In:
  Proceedings of the IEEE/CVF Conference on Computer Vision and Pattern
  Recognition. pp. 1290--1299 (2022)

\bibitem{robust_learning_image_quality_marine_waste}
Cheng, Y., Zhu, J., Jiang, M., Fu, J., Pang, C., Wang, P., Sankaran, K.,
  Onabola, O., Liu, Y., Liu, D., et~al.: Flow: A dataset and benchmark for
  floating waste detection in inland waters. In: Proceedings of the IEEE/CVF
  International Conference on Computer Vision. pp. 10953--10962 (2021)

\bibitem{dawid_skene}
Dawid, A.P., Skene, A.M.: Maximum likelihood estimation of observer error-rates
  using the em algorithm. Journal of the Royal Statistical Society: Series C
  (Applied Statistics)  \textbf{28}(1),  20--28 (1979)

\bibitem{robust_learning_jiou}
Feng, D., Wang, Z., Zhou, Y., Rosenbaum, L., Timm, F., Dietmayer, K., Tomizuka,
  M., Zhan, W.: Labels are not perfect: Inferring spatial uncertainty in object
  detection. IEEE Transactions on Intelligent Transportation Systems  (2021)

\bibitem{robust_learning_noise_tolerant}
Gao, J., Wang, J., Dai, S., Li, L.J., Nevatia, R.: Note-rcnn: Noise tolerant
  ensemble rcnn for semi-supervised object detection. In: Proceedings of the
  IEEE/CVF international conference on computer vision. pp. 9508--9517 (2019)

\bibitem{gao_data_dependent}
Gao, Z., Sun, F.K., Yang, M., Ren, S., Xiong, Z., Engeler, M., Burazer, A.,
  Wildling, L., Daniel, L., Boning, D.S.: Learning from multiple annotator
  noisy labels via sample-wise label fusion. In: Computer Vision--ECCV 2022:
  17th European Conference, Tel Aviv, Israel, October 23--27, 2022,
  Proceedings, Part XXIV. pp. 407--422. Springer (2022)

\bibitem{wdn}
Guan, M., Gulshan, V., Dai, A., Hinton, G.: Who said what: Modeling individual
  labelers improves classification. In: Proceedings of the AAAI conference on
  artificial intelligence. vol.~32 (2018)

\bibitem{mask_rcnn}
He, K., Gkioxari, G., Doll{\'a}r, P., Girshick, R.: Mask r-cnn. In: Proceedings
  of the IEEE international conference on computer vision. pp. 2961--2969
  (2017)

\bibitem{dl_noisy_labels_medical}
Karimi, D., Dou, H., Warfield, S.K., Gholipour, A.: Deep learning with noisy
  labels: Exploring techniques and remedies in medical image analysis. Medical
  image analysis  \textbf{65},  101759 (2020)

\bibitem{mbem}
Khetan, A., Lipton, Z.C., Anandkumar, A.: Learning from noisy singly-labeled
  data. arXiv preprint arXiv:1712.04577  (2017)

\bibitem{robust_learning_domain_shift_3}
Khodabandeh, M., Vahdat, A., Ranjbar, M., Macready, W.G.: A robust learning
  approach to domain adaptive object detection. In: Proceedings of the IEEE/CVF
  International Conference on Computer Vision. pp. 480--490 (2019)

\bibitem{simple_method}
Langerak, T.R., van~der Heide, U.A., Kotte, A.N., Viergever, M.A., Van~Vulpen,
  M., Pluim, J.P.: Label fusion in atlas-based segmentation using a selective
  and iterative method for performance level estimation (simple). IEEE
  transactions on medical imaging  \textbf{29}(12),  2000--2008 (2010)

\bibitem{vindr_dataset}
Le, K.H., Tran, T.V., Pham, H.H., Nguyen, H.T., Le, T.T., Nguyen, H.Q.:
  Learning from multiple expert annotators for enhancing anomaly detection in
  medical image analysis. arXiv preprint arXiv:2203.10611  (2022)

\bibitem{docbank}
Li, M., Xu, Y., Cui, L., Huang, S., Wei, F., Li, Z., Zhou, M.: Docbank: A
  benchmark dataset for document layout analysis. arXiv preprint
  arXiv:2006.01038  (2020)

\bibitem{coco}
Lin, T.Y., Maire, M., Belongie, S., Hays, J., Perona, P., Ramanan, D.,
  Doll{\'a}r, P., Zitnick, C.L.: Microsoft coco: Common objects in context. In:
  European conference on computer vision. pp. 740--755. Springer (2014)

\bibitem{robust_learning_image_quality_datasets}
Michaelis, C., Mitzkus, B., Geirhos, R., Rusak, E., Bringmann, O., Ecker, A.S.,
  Bethge, M., Brendel, W.: Benchmarking robustness in object detection:
  Autonomous driving when winter is coming. arXiv preprint arXiv:1907.07484
  (2019)

\bibitem{vinbigdata-kaggle}
Nguyen, D.B., Nguyen, H.Q., Elliott, J., KeepLearning, Nguyen, N.T., Culliton,
  P.: Vinbigdata chest x-ray abnormalities detection (2020),
  \url{https://kaggle.com/competitions/vinbigdata-chest-xray-abnormalities-detection}

\bibitem{nguyen2022vindr}
Nguyen, H.Q., Lam, K., Le, L.T., Pham, H.H., Tran, D.Q., Nguyen, D.B., Le,
  D.D., Pham, C.M., Tong, H.T., Dinh, D.H., et~al.: Vindr-cxr: An open dataset
  of chest x-rays with radiologist’s annotations. Scientific Data
  \textbf{9}(1), ~429 (2022)

\bibitem{detecto_rs}
Qiao, S., Chen, L.C., Yuille, A.: Detectors: Detecting objects with recursive
  feature pyramid and switchable atrous convolution. In: Proceedings of the
  IEEE/CVF conference on computer vision and pattern recognition. pp.
  10213--10224 (2021)

\bibitem{robust_learning_domain_shift}
Ramamonjison, R., Banitalebi-Dehkordi, A., Kang, X., Bai, X., Zhang, Y.:
  {SimROD}: {A} {Simple} {Adaptation} {Method} for {Robust} {Object}
  {Detection}. In: 2021 {IEEE}/{CVF} {International} {Conference} on {Computer}
  {Vision} ({ICCV}). pp. 3550--3559. IEEE, Montreal, QC, Canada (Oct 2021).
  \doi{10.1109/ICCV48922.2021.00355},
  \url{https://ieeexplore.ieee.org/document/9711168/}

\bibitem{mjv_2}
Raykar, V.C., Yu, S., Zhao, L.H., Jerebko, A., Florin, C., Valadez, G.H.,
  Bogoni, L., Moy, L.: Supervised learning from multiple experts: whom to trust
  when everyone lies a bit. In: Proceedings of the 26th Annual international
  conference on machine learning. pp. 889--896 (2009)

\bibitem{faster_rcnn}
Ren, S., He, K., Girshick, R., Sun, J.: Faster r-cnn: Towards real-time object
  detection with region proposal networks. Advances in neural information
  processing systems  \textbf{28} (2015)

\bibitem{dl_f_crowds}
Rodrigues, F., Pereira, F.: Deep learning from crowds. In: Proceedings of the
  AAAI conference on artificial intelligence. vol.~32 (2018)

\bibitem{mjv_1}
Sheng, V.S., Provost, F., Ipeirotis, P.G.: Get another label? improving data
  quality and data mining using multiple, noisy labelers. In: Proceedings of
  the 14th ACM SIGKDD international conference on Knowledge discovery and data
  mining. pp. 614--622 (2008)

\bibitem{ml_crowdsourcing}
Sheng, V.S., Zhang, J.: Machine learning with crowdsourcing: A brief summary of
  the past research and future directions. In: Proceedings of the AAAI
  conference on artificial intelligence. vol.~33, pp. 9837--9843 (2019)

\bibitem{fast_dawid_skene}
Sinha, V.B., Rao, S., Balasubramanian, V.N.: Fast dawid-skene: A fast vote
  aggregation scheme for sentiment classification. arXiv preprint
  arXiv:1803.02781  (2018)

\bibitem{wbf}
Solovyev, R., Wang, W., Gabruseva, T.: Weighted boxes fusion: Ensembling boxes
  from different object detection models. Image and Vision Computing
  \textbf{107},  104117 (2021)

\bibitem{survey_noisy_labels}
Song, H., Kim, M., Park, D., Shin, Y., Lee, J.G.: Learning from noisy labels
  with deep neural networks: A survey. IEEE Transactions on Neural Networks and
  Learning Systems  (2022)

\bibitem{tanno2019learning}
Tanno, R., Saeedi, A., Sankaranarayanan, S., Alexander, D.C., Silberman, N.:
  Learning from noisy labels by regularized estimation of annotator confusion.
  In: Proceedings of the IEEE/CVF conference on computer vision and pattern
  recognition. pp. 11244--11253 (2019)

\bibitem{own_layout_paper}
Tschirschwitz, D., Klemstein, F., Stein, B., Rodehorst, V.: A dataset for
  analysing complex document layouts in the digital humanities and its
  evaluation with krippendorff’s alpha. In: DAGM German Conference on Pattern
  Recognition. pp. 354--374. Springer (2022)

\bibitem{robust_learning_domain_shift_2}
Wang, X., Huang, T.E., Liu, B., Yu, F., Wang, X., Gonzalez, J.E., Darrell, T.:
  Robust object detection via instance-level temporal cycle confusion. In:
  Proceedings of the IEEE/CVF International Conference on Computer Vision. pp.
  9143--9152 (2021)

\bibitem{robust_learning_pseudo_labels}
Wang, Z., Li, Y., Guo, Y., Fang, L., Wang, S.: Data-uncertainty guided
  multi-phase learning for semi-supervised object detection. In: Proceedings of
  the IEEE/CVF Conference on Computer Vision and Pattern Recognition. pp.
  4568--4577 (2021)

\bibitem{staple_method}
Warfield, S.K., Zou, K.H., Wells, W.M.: Simultaneous truth and performance
  level estimation (staple): an algorithm for the validation of image
  segmentation. IEEE transactions on medical imaging  \textbf{23}(7),  903--921
  (2004)

\bibitem{glad}
Whitehill, J., Wu, T.f., Bergsma, J., Movellan, J., Ruvolo, P.: Whose vote
  should count more: Optimal integration of labels from labelers of unknown
  expertise. Advances in neural information processing systems  \textbf{22}
  (2009)

\bibitem{double_head}
Wu, Y., Chen, Y., Yuan, L., Liu, Z., Wang, L., Li, H., Fu, Y.: Rethinking
  classification and localization for object detection. In: Proceedings of the
  IEEE/CVF conference on computer vision and pattern recognition. pp.
  10186--10195 (2020)

\bibitem{robust_learning_image_quality_drones}
Wu, Z., Suresh, K., Narayanan, P., Xu, H., Kwon, H., Wang, Z.: Delving into
  robust object detection from unmanned aerial vehicles: A deep nuisance
  disentanglement approach. In: Proceedings of the IEEE/CVF International
  Conference on Computer Vision. pp. 1201--1210 (2019)

\bibitem{robust_learning_adversarial_2}
Xie, C., Wang, J., Zhang, Z., Zhou, Y., Xie, L., Yuille, A.: Adversarial
  examples for semantic segmentation and object detection. In: Proceedings of
  the IEEE international conference on computer vision. pp. 1369--1378 (2017)

\bibitem{robust_learning_adversarial}
Zhang, H., Wang, J.: Towards {Adversarially} {Robust} {Object} {Detection}. In:
  2019 {IEEE}/{CVF} {International} {Conference} on {Computer} {Vision}
  ({ICCV}). pp. 421--430. IEEE, Seoul, Korea (South) (Oct 2019).
  \doi{10.1109/ICCV.2019.00051},
  \url{https://ieeexplore.ieee.org/document/9009990/}

\bibitem{distilling_severe_label_noise}
Zhang, Z., Zhang, H., Arik, S.O., Lee, H., Pfister, T.: Distilling effective
  supervision from severe label noise. In: Proceedings of the IEEE/CVF
  Conference on Computer Vision and Pattern Recognition. pp. 9294--9303 (2020)

\bibitem{crowdsourcing_GTI}
Zheng, Y., Li, G., Li, Y., Shan, C., Cheng, R.: Truth inference in
  crowdsourcing: Is the problem solved? Proceedings of the VLDB Endowment
  \textbf{10}(5),  541--552 (2017)

\bibitem{publaynet}
Zhong, X., Tang, J., Yepes, A.J.: Publaynet: largest dataset ever for document
  layout analysis. In: 2019 International Conference on Document Analysis and
  Recognition (ICDAR). pp. 1015--1022. IEEE (2019)

\bibitem{noise_types}
Zhu, X., Wu, X.: Class noise vs. attribute noise: A quantitative study. The
  Artificial Intelligence Review  \textbf{22}(3), ~177 (2004)

\end{thebibliography}

\newpage

\section*{Appendix 1}
\label{appx_1}
\label{sec:wmf}

This section presents our adaptation of the weighted box fusion (WBF) technique, tailored specifically for instance segmentation as a weighted mask fusion (WMF).

In their study, \cite{vindr_dataset} propose a method for combining annotations from multiple annotators using a weighted box fusion \cite{wbf} approach. In this method, bounding boxes are matched greedily only with boxes of the same class, and no annotations are discarded. The WBF algorithm fuses boxes that exceed a specified overlap threshold, resulting in new boxes that represent the weighted average of the original boxes. The approach also allows for inclusion of box confidence scores and prior weights for each annotator.

To extend the WBF method for instance segmentation, we introduce an option to fuse segmentation masks, which involves four steps: (1) calculating the weighted area and weighted center points from the different masks, (2) compute the average center point and average area from the selected masks, (3) determining the closest center point of the original masks to the weighted center point and selecting this mask, and (4) dilating or eroding the chosen mask until the area is close to the averaged area. The resulting mask is used as the aggregated segmentation mask and is also used as the averaging operation during the aggregation for LAEM and MJV with uniform weight.

Moreover, we integrate the WBF approach with LAEM, yielding WBF+EM. This integration involves assessing annotator confidence using LAEM, and subsequently incorporating it into the WBF method to produce weighted average areas instead of simply averaged areas. While the differences between LAEM and WBF might seem subtle, WBF+EM offers a more thorough approach to annotator fusion. This modification is relatively minor, and its impact is modest, as corroborated by our experiments delineated in \nameref{appx_2}.

\section*{Appendix 2}
\label{appx_2}

% \usepackage{multirow}
\begin{table}[]
\newcolumntype{N}{>{\centering\arraybackslash}p{7ex}}
\centering
\begin{tabular}{c|ccl|N|NNN|NNN|NNN}
\toprule
\multicolumn{4}{c|}{DetectoRS} & \multicolumn{9}{c}{Training} \\ \cline{5-13} 
\multicolumn{4}{c|}{\multirow{2}{*}{TexBiG}} & \multicolumn{1}{c|}{\multirow{2}{*}{RL}} & \multicolumn{3}{c|}{MJV} & \multicolumn{3}{c|}{LAEM} & \multicolumn{2}{c}{WBF} \\
\multicolumn{4}{c|}{} & \multicolumn{1}{l|}{} & \multicolumn{1}{c}{$\cup$} & \multicolumn{1}{c}{$\mu$} & \multicolumn{1}{c|}{$\cap$} & \multicolumn{1}{c}{$\cup$} & \multicolumn{1}{c}{$\mu$} & \multicolumn{1}{c|}{$\cap$} & \multicolumn{1}{c}{base} & \multicolumn{1}{c}{EM} \\ \hline
\multirow{16}{*}{\rotatebox[origin=c]{90}{Test}} & \multirow{6}{*}{MJV}  & $\cup$  & AP   & 32.5 & 34.5 & 30.4 & 25.5 & \textbf{35.1} & 29.1 & 28.4 & 30.2 & 30.7 \\
 & & & AP$^{bb}$ & 34.7 & 36.6 & 34.0 & 29.9 & \textbf{37.5} & 32.7 & 33.4 & 34.4 & 33.5  \\
 & & \multirow{2}{*}{$\mu$} & AP & 41.9 & \textbf{43.9} & 39.9 & 35.8 & 43.6 & 40.8 & 35.0 & 41.5 & 41.9  \\
 & & & AP$^{bb}$ & 45.6 & \textbf{48.2} & 44.6 & 41.4 & 47.9                      & 44.5                          & 40.2         & 45.5                      & 46.2                   \\
                       &                       & \multirow{2}{*}{$\cap$} & AP   & 44.2                             & 41.7                      & \textbf{46.6}                          & 45.0         & 43.2                      & 45.3                          & 45.2         & 46.0                      & 44.9                   \\
                       &                       &                               & AP$^{bb}$ & 49.6                            & 47.9                      & \textbf{51.4}                          & 49.9         & 49.3                      & 50.6                          & 49.5         & 51.0                      & 49.2                   \\ \cline{2-13} 
                       & \multirow{6}{*}{\rotatebox[origin=c]{90}{LAEM}} & \multirow{2}{*}{$\cup$}        & AP   & 31.5                             & 34.9                      & \textbf{44.4}                          & 25.9         & 33.6                      & 30.5                          & 26.8         & 29.8                      & 31.1                   \\
                       &                       &                               & AP$^{bb}$ & 33.6                             & 36.5                      & \textbf{48.9}                          & 30.5         & 35.8                      & 33.8                          & 31.4         & 33.0                      & 34.6                   \\
                       &                       & \multirow{2}{*}{$\mu$}    & AP   & 41.1                             & 42.5                      & 40.5                          & 34.9         & \textbf{43.6}                      & 40.2                          & 35.9         & 40.7                      & 40.4                   \\
                       &                       &                               & AP$^{bb}$ & 44.8                             & 46.8                      & 44.6                          & 41.9         & \textbf{47.7}                      & 44.3                          & 41.7         & 45.5                      & 45.1                   \\
                       &                       & \multirow{2}{*}{$\cap$} & AP   & 43.5                             & 40.8                      & 43.0                          & 45.0         & 41.6                      & 44.1                          & 44.0         & 45.0                      & \textbf{45.1}                   \\
                       &                       &                               & AP$^{bb}$ & 49.8                             & 46.0                      & 48.5                          & 49.5         & 46.9                      & 48.9                          & 48.0         & 49.5                      &  \textbf{50.3}                   \\ \cline{2-13} 
                       & \multirow{4}{*}{\rotatebox[origin=c]{90}{WBF}}  & \multirow{2}{*}{base}        & AP   & 36.1                             & \textbf{38.0}                      & 34.8                          & 33.4         & 37.3                      & 30.3                          & 32.7         & 34.9                      & 36.9                   \\
                       &                       &                               & AP$^{bb}$ & 38.8                             & \textbf{41.6}                      & 38.0                          & 37.4         & 40.2                      & 33.6                          & 37.8         & 38.5                      & 40.3                   \\
                       &                       & \multirow{2}{*}{EM}           & AP   & 38.1                             & \textbf{39.9}                      & 34.9                          & 32.4        & 39.8                      & 36.5                          & 32.8         & 36.3                      & 35.9                   \\
                       &                       &                               & AP$^{bb}$ & 40.6                             & \textbf{42.8}                      & 38.3                          & 36.4         & 42.8                      & 40.2                          & 37.6         & 40.0                      & 39.3                   \\ \hline
\multicolumn{3}{c}{\multirow{2}{*}{Mean}}                                      & AP   & 38.6                             & 39.5                      & 39.3                          & 34.7         & \textbf{39.7}                      & 37.1                          & 35.1         & 38.1                      & 38.4                   \\
\multicolumn{3}{l}{}                                                           & AP$^{bb}$ & 42.2                             & 43.3                      & \textbf{43.5}                          & 39.6         & \textbf{43.5}                      & 41.1                          & 40.0         & 42.2                      & 42.3  \\
\bottomrule
\end{tabular}
\caption{Cross-Validation of ground truth inference combinations between training and test data, for the DetectoRS with a ResNet-50 backbone on the TexBiG dataset. Showing the $mAP@[.5:.95]$ for instance masks and bounding boxes. Union is represented by $\cup$, intersection by $\cap$ and averaging by $\mu$. RL denotes training conducted on un-aggregated noisy labels. The two rows on the bottom show how the training methods perform on average.}
\label{cross_val_texbig}
\end{table}

In this experiment, we carried out a comparative analysis of different ground truth inference methods. To do this, we separated the annotations for training and testing, and created various combinations of train-test datasets using the available ground truth estimation methods. Afterward, a model was trained on these combinations. The results from this experiment reveal how aggregation methods can impact the performance of the trained models and show how these outcomes can vary based on the specific combination of training and testing aggregation used.

Tables \ref{cross_val_texbig} and \ref{cross_val_vindrcxr} present the application of various ground truth estimation methods on repeated labels. In the TexBiG dataset, each method is employed to aggregate the labels of both training and test data, and all possible train-test combinations are learned and tested to perform a cross comparison of the different ground truth inference methods, as shown in Table \ref{cross_val_texbig}. The hyperparameter for the area combination is denoted as $\cup$ for union, $\mu$ for averaging and $\cap$ for intersection. Additionally, the plain repeated labels, without any aggregation, are compared with the different aggregated test data. Our findings reveal that on a high-agreement dataset, weighted boxes fusion does not perform well. This could be attributed to the inclusion of most annotations by WBF, whereas in cases with high agreement, it is more desirable to exclude non-conforming instances. Majority voting and localization-aware expectation maximization perform similarly; however, LAEM provides a more elegant solution for addressing edge cases. Calculating the annotator confidence, as performed in LAEM, is highly advantageous. However, in rare cases, spammer annotators could potentially circumvent annotation confidence by annotating large portions of simple examples correctly but failing at hard cases. Such cases would result in a high confidence level for the spammer, potentially outvoting the correct annotators on challenging and crucial cases.

% \usepackage{multirow}
\begin{table}[]
\centering
\newcolumntype{N}{>{\centering\arraybackslash}p{6ex}}
\begin{tabular}{c|N|NNN|NNN|NN}
\toprule
\multirow{3}{*}{\begin{tabular}[c]{@{}c@{}}Double Head R-CNN\\ VinDr-CXR\end{tabular}} & \multicolumn{9}{c}{Training}                                                                                                                           \\ \cline{2-10} 
                                                                                       & \multicolumn{1}{c|}{\multirow{2}{*}{RL}} & \multicolumn{3}{c|}{MJV}                & \multicolumn{3}{c|}{LAEM}               & \multicolumn{2}{c}{WBF} \\
                                                                                       & \multicolumn{1}{c|}{}                    & $\cup$    & $\mu$    & \multicolumn{1}{c|}{$\cap$}    & $\cup$    & $\mu$    & \multicolumn{1}{c|}{$\cap$}    & base       & EM         \\ \hline
private LB                                                                             & \textbf{16.2}                & 15.2 & 14.6 & 14.3 & 14.9 & 15.0 & 14.7 & 13.7       & 14.3      \\
\bottomrule
\end{tabular}
\caption{Comparing results with the private Kaggle leaderboard \cite{vinbigdata-kaggle} for the VinDr-CXR dataset using the double headed R-CNN at $mAP_{40}$. Union is represented by $\cup$, intersection by $\cap$ and averaging by $\mu$. RL denotes training conducted on un-aggregated noisy labels.}
\label{cross_val_vindrcxr}
\end{table}

The main performance differences between MJV and LAEM arise due to the application of the three different combination operations -- union, averaging, and intersection. Combining areas by taking their union results in larger areas, making it easier for a classifier to identify the respective regions. Analysis of the mean results of the training methods reveals that both MJV+$\cup$ and LAEM+$\cup$ exhibit the highest performance across various test configurations. On the contrary, methods parameterized with intersection $\cap$ yield the lowest mean results. Training with repeated labels without any aggregation yields results similar to training with aggregated labels.  However, while it may be generally feasible to train with noisy labels, the performance is slightly dampened. Since the test data aggregation method is LAEM-$\mu$ as described in Section \ref{algo_design}, the best performing training method LAEM-$\cup$ is choosen as the aggregation method for the training data in the experiments shown in Section \ref{contingent_study} and \ref{leave_out_study}.

For the VinDr-CXR dataset, a smaller, similar experiment is performed as shown in Table \ref{cross_val_vindrcxr}. As the Kaggle leaderboard already provides an aggregated ground truth and labels are unavailable, only the training data are aggregated. Our findings indicate that training with plain repeated labels leads to higher results. Given the low agreement of the dataset, training with repeated labels may be seen as a form of ``label augmentation.'' Interestingly, the methods used to aggregate the test data, such as WBF, do not outperform the other methods. However, ground truth estimation methods are not designed to boost performance but rather to provide a suitable estimation for the targeted outcome. Based on these results, the following experiments on VinDr-CXR will be run with the repeated labels for training.

\section*{Appendix 3}
\label{appx_3}

This section shows three more comparisons between different ground truth aggregation methods, exemplary on the VinDr-CXR dataset \cite{nguyen2022vindr}. All of them follow the same structure. Left: the original image with the repeated labels indicated by the different line types. Right: the four smaller images from top left to bottom right are, MJV+$\cap$, LAEM+$\mu$, LAEM+$\cup$ and WBF.

\begin{figure}[t]
    \centering
    \def\figurewidth{0.85}
    \includegraphics[width=\figurewidth\linewidth]{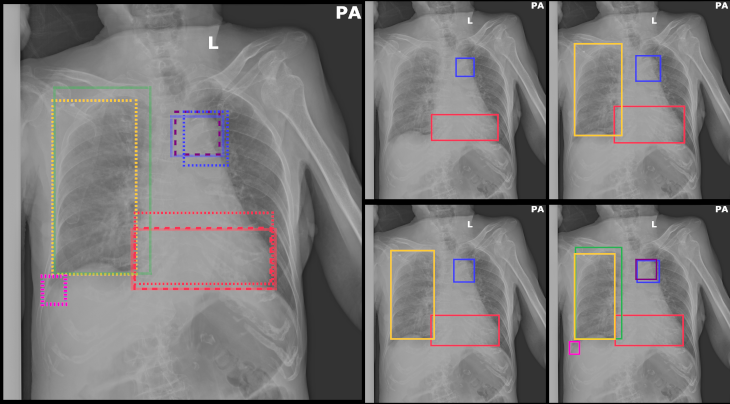}
    \includegraphics[width=\figurewidth\linewidth]{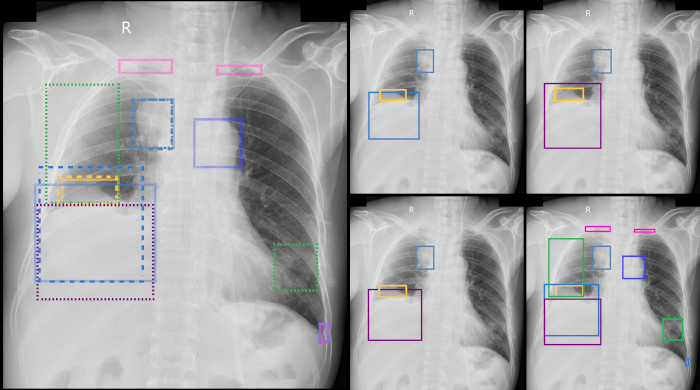}
    \includegraphics[width=\figurewidth\linewidth]{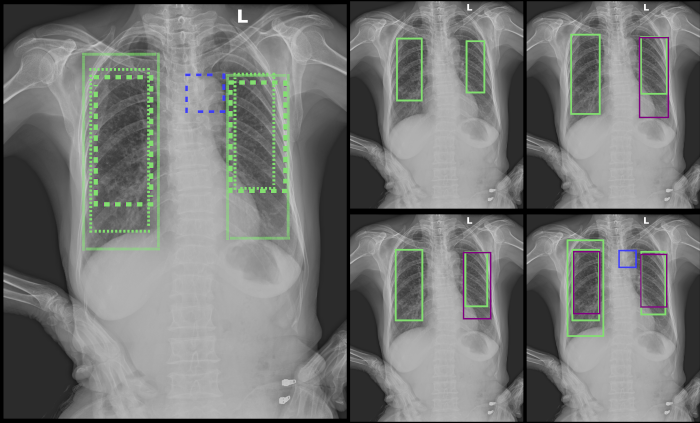}
    %\vspace{-0.5cm}
    \caption{Qualitative results on three test images from the VinDr-CXR. Left: the original image with the repeated labels indicated by the different line types. Right: the four smaller images from top left to bottom right are, MJV+$\cap$, LAEM+$\mu$, LAEM+$\cup$ and WBF. }
\end{figure}

\end{document}